\documentclass{llncs}
\usepackage{graphicx}
\usepackage{lscape}
\usepackage{amsmath}
\usepackage{subcaption}
\usepackage{mathtools}
\usepackage{float}
\usepackage{tabularx}

\begin{document}
\bibliographystyle{./splncs}
\title{Towards safe deep learning: accurately quantifying biomarker uncertainty in neural network predictions}

\author{  Zach Eaton-Rosen\inst{1} 
	\and Felix Bragman\inst{1}	
	\and Sotirios Bisdas\inst{2, 3}
	\and Sebastien Ourselin\inst{4}
	\and M. Jorge Cardoso \inst{1, 4} }



\institute{$^1$Centre for Medical Image Computing, University College London, UK\\
	$^2$Department of Neuroradiology, The National Hospital for Neurology and Neurosurgery, University College London NHS Foundation Trust, UK\\
	$^3$
	Department of Brain Repair and Rehabilitation, Institute of Neurology, University College London, UK\\
	$^4$Biomedical Engineering and Imaging Sciences, Kings College London, UK}

\maketitle

\begin{abstract}  

Automated medical image segmentation, specifically using deep learning, has shown outstanding performance in semantic segmentation tasks. However, these methods rarely quantify their uncertainty, which may lead to errors in downstream analysis. In this work we propose to use Bayesian neural networks to quantify uncertainty within the domain of semantic segmentation. We also propose a method to convert voxel-wise segmentation uncertainty into volumetric uncertainty, and calibrate the accuracy and reliability of confidence intervals of derived measurements. When applied to a tumour volume estimation application, we demonstrate that by using such modelling of uncertainty, deep learning systems can be made to report volume estimates with well-calibrated error-bars, making them safer for clinical use. We also show that the uncertainty estimates extrapolate to unseen data, and that the confidence intervals are robust in the presence of artificial noise. This could be used to provide a form of quality control and quality assurance, and may permit further adoption of deep learning tools in the clinic.

\end{abstract}

\section{Introduction}   


Deep convolutional neural nets (CNNs) are becoming the dominant method for medical image semantic segmentation, markedly improving on previous techniques in a variety of tasks. Standard deep learning models produce a point-estimate of the `probability' that a voxel belongs to each segmentation class --- but these estimates typically lack any quantification of error. In medical applications safety is paramount: misreporting a biomarker or outright missing a pathological finding may endanger patients. If uncertainty estimates are not propagated through a clinical pipeline, or if errors are not properly quantified and calibrated, this can result in false conclusions downstream, e.g. when comparing volume estimates longitudinally. Models that can quantify uncertainty in their predictions may thus be useful in making safer and more actionable analysis pipelines. 

While it remains uncommon to calculate estimates of uncertainty in probabilistic segmentation -- indeed, many methods preclude it entirely -- Bayesian models of segmentation incorporate estimates of model parameter variance explicitly, and even downstream errors such as volume estimates~\cite{iglesias2013improved}. These methods have however seen limited adoption: they are computationally expensive, especially compared with modern CNNs, and have worse accuracy. In addition, the formulation in~\cite{iglesias2013improved} relies on having a probabilistic atlas to use as a prior term, which may not be amenable to highly variable pathological tissues such as tumours or Multiple Sclerosis lesions. 
Bayesian neural networks can be used to approximate model uncertainty even without probabilistic atlases. One method of approximating a Bayesian neural network is to use `dropout', which was originally proposed as a regularising technique~\cite{srivastava2014dropout}. 
In dropout, a randomly-determined subset of neurons in the network are `dropped' randomly (have zero output) at each iteration. With modern neural networks having upwards of millions of parameters, this means that the same dropout mask is unlikely to ever be chosen twice for the network. At each iteration, one essentially uses a specific, thinned, version of the network. The stochastic nature of these networks can be used to approximate a Bayesian neural network~\cite{gal2016dropout}. 
From a practical point of view, predictive uncertainty is estimated by calculating the sample variance of predictions made from different forward passes from the network. 


Bayesian modeling of uncertainties in CNNs 
is still not commonplace. To the best of our knowledge, no prior work has made use of these uncertainties to estimate the impact and reliability of downstream biomarker uncertainty (e.g. volume estimated from a segmentation), nor to describe how uncertainty estimates are related to data quality. In this work, we utilise a Bayesian deep learning model to measure the uncertainty of image segmentation. We fit network architectures with differing levels of stochasticity. We then measure the effects of uncertainty on volume estimation and propose to use these stochastic network outputs to build `contours' of increasing segmentation volume that better approximate the unknown volume uncertainties. 
Because these techniques can be applied to any neural networks trained with dropout, our proposed method can be employed to give calibrated estimates for volume measurements at little additional cost.

\section{Methods}\label{sec:methods}

In the Brain Tumour Segmentation (BraTS) challenge~\cite{menze2015multimodal,bakas2017advancing}, CNNs have been used to produce winning submissions in recent years and have been established as the dominant fully-automated method for the task. We use data from the BraTS 2017 training dataset which contains 285 subjects with high- or low-grade gliomas. Each subject has T$_1$1-weighted, T$_1$-weighted with Gd contrast (T$_1$ce), T$_2$-weighted and FLAIR MRI scans. One expert segmentation is given for each subject, with 3 foreground labels: (1) Gd-enhancing tumour, (2) edema, and (3) the necrotic/non-enhancing tumour. These are combined to get binary hierarchical segmentations of `active'(1) , `core' (1 + 3) and `whole' (1 + 2 + 3), which we use henceforth. 

This work concentrates on introducing a generalisable technique for quantifying uncertainty in a network's outputs rather than on designing a new neural network architecture. To that end, we used multiple variants of `High-Res Net'~\cite{li2017compactness}, a residual network that uses dilated convolutions to increase the effective receptive field and the context of the representation. The proposed architectures are designed to test the effects of different levels of uncertainty throughout the network. The dropout rate $p$ is kept constant at 0.5 except if applied in the input layer, in which case it is set to 0.05 (a low value is chosen to allow low-level image features into the network). The value of 0.5 maximises the variance of the layer outputs, which can be seen as maximising the effect of dropout.  
The number of kernels is larger than in \cite{li2017compactness} to allow the network to cope with the reduced capacity caused by dropout. We also train a non-stochastic (no dropout) network with the same number of parameters as a baseline.

In total, we trained four variants of the High-Res Net architecture: all networks use filter sizes of \{24, 24, 48, 96, 96\} for the first convolutional layer, the three convolutional blocks, and the next convolutional layer respectively. We define: $HR_{\text{default}}$: a non-stochastic network (no dropout); HR$_{drop\_last}$: a stochastic network with dropout in the last layer; HR$_{drop\_all}$: a stochastic network with dropout after each residual block, and a stochastic network with a heteroscedastic noise model HR$_{hetero}$. The heteroscedastic model uses the same network architecture as a trunk that branches into (a): a convolution with kernel of dimension $3\times3\times3$ and 120 filters before the softmax layer (for segmentation) and (b): a convolutional layer of dimension $3\times3\times3$ and 80 filters that connects to a softplus layer to output the estimate of the standard deviation. 
 
`Predictive variance' denotes variance obtained by running these models several times and calculating variance over multiple trials. The heteroscedastic variant, HR$_{hetero}$, also adds an additive `aleatoric' variance, which quantifies uncertainty over the data itself. 

All optimisation is performed with the Adam optimiser with learning rate 0.001. All results are reported on withheld `test' data with a 70:20:10 train: validation:~test~split. Training is halted when the validation loss does not improve for over 5 epochs. The work is implemented using NiftyNet \cite{GIBSON2018113}, and code will be made publicly available. 

\subsection{Bayesian Deep Learning}

In Bayesian neural networks, model weights \textbf{W} are assumed to have a distribution, rather than being point estimates. We aim to approximate the posterior $p$(\textbf{W}$|$\textbf{X},\textbf{Y}) over the weights \textbf{W} given training data $\{\text{\textbf{X}},\text{\textbf{Y}}\}$. Dropout samples from the space of sub-models of a network architecture, where sampling is parameterised by a randomly-sampled Bernoulli dropout mask, to ultimately estimate model parameters and their uncertainties. Using this formulation, the neural network approximates a Gaussian Process~\cite{gal2016dropout}.

In the proposed networks, we predict probabilities for a voxel to belong to each segmentation class. At test-time, we output the model $T$ times, with $T$ different dropout masks, and use these predictions to estimate the uncertainty. Here, $y$ is the prediction of the network and $\hat{y}$ is the result of one stochastic approximation. This variance can then be given explicitly by $\text{Var}_{epi}(y) \approx \frac{1}{T} \sum_{t=1}^{T} \hat{y}^2 - \left(\frac{1}{T} \sum_{t=1}^{T} \hat{y}\right)^2$, i.e. estimated by measuring the variance of the forward passes. For networks except HR$_{hetero}$, we use a cross-entropy loss function. 

This variance takes into account the uncertainty in the model itself by using different dropout masks for differing predictions. However, other sources of variance can also be measured. As in~\cite{kendall2017uncertainties}, an additional term can be introduced to model the uncertainty that is intrinsic to the problem --- known as \textit{aleatoric} uncertainty. With this new term, the variance can now be expressed as: $\text{Var}(y) \approx \text{Var}_{epistemic}(y) + \frac{1}{T}\sum_{t=1}^{T} \sigma_t ^ 2 $. 
While this aleatoric uncertainty can be approximated in many ways, for numerical stability, we follow~\cite{kendall2017uncertainties} in assuming that the error has a normal distribution in logit space (un-normalised probabilities), and add the aleatoric noise to the logits directly in the network: $\hat{x}_{i,t} = f_{i}^W + \sigma_i^W\epsilon_t, \epsilon_t \sim \mathcal{N}(0, 1)$. 
This noise model can then be used in an heteroscedastic setup through the following loss function $\mathcal{L}_{x} = \frac{1}{N} \sum_{i}^{N} \log\frac{1}{T}\sum_{t} \exp(\hat{x}_{i,t,c} - \log\sum_{c'}\exp \hat{x}_{i,t,c'})$. Note that this loss requires multiple passes over several noise realisations, and that increasing the noise in a voxel will reduce the logit differences and, correspondingly, the certainty of the predictions.

\subsection{From stochastic segmentation samples to calibrated volumetric uncertainty}
\label{sec:seguncert}

The uncertainty in these models can be visualised directly in the segmentation space, but it is harder to propagate this uncertainty from pixel-wise segmentation to volume estimation. 

For a given subject $i$, the foreground volume $V = \sum_{i=1}^{n}{p(x_i)}$. Combining errors from the $i$ voxels, we get: $Var(V) = \sum_{i=1}^{n}\sum_{j=1}^{n}{cov(x_i, x_j)}$. 
Although the covariance terms could be measured empirically , we instead decided to explicitly spatially regularise the data. 
We did this by predicting the probability of being in the foreground class repeatedly. We then take quantiles of this measurement at each voxel to build a cumulative distribution. In terms of `combining errors', this is the equivalent of maximising the correlation between all voxels. 

Maximising this correlation between voxels' predictions is not guaranteed to \emph{calibrate} the volumetric estimates. To do this, we introduce a step to fit the error bars for the validation data. 
Practically, this is achieved by using the validation to fit an affine transformation on the percentiles of the volumetric CDF to a uniform distribution --- equivalent to a 1-D histogram equalization. This mapping enforces the correct proportion of ground truths to appear in a given confidence interval. For final results, the parameters of the scaling for the `test' set are fitted on all the `validation' data. In the `validation' data, to avoid test-train contamination, the parameters of the fit are determined through a 3-fold paradigm.

\section{Results}

\begin{figure}[t]
	\centering
	\includegraphics[width=1.\linewidth, trim={0 1.5cm 0 .5cm}, clip]{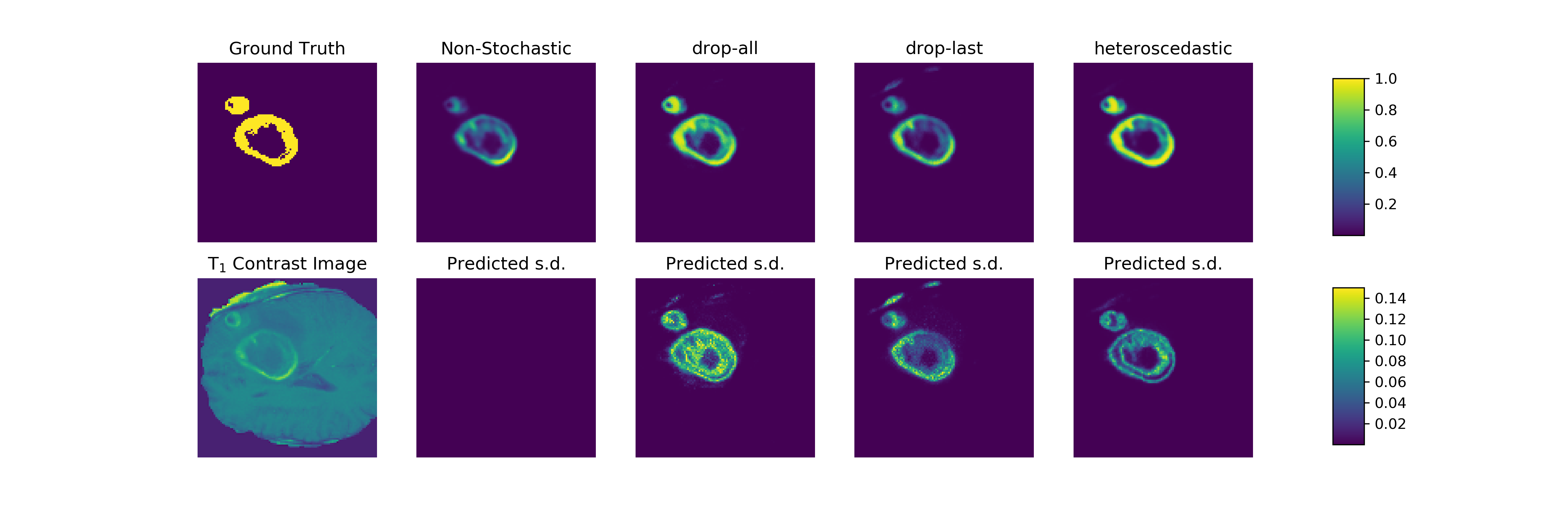}
	\caption{Top row: mean confidences (over 20 forward passes) for the given model to belong to the `Active' class. Bottom row: $\sigma_{predicted}$, the standard deviation across predictions. Different methods display similar predictions, but the level of uncertainty varies depending on the network used.}
	\label{fig:brains}
\end{figure}

\begin{figure}[b]
	\centering
	\includegraphics[width=.8\linewidth, trim={0 0cm 0 1cm}, clip]{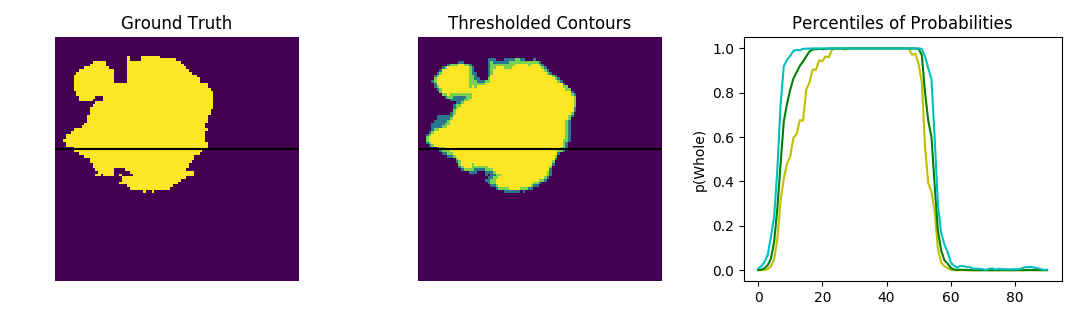}
	\caption{Left: ground truth `Whole' segmentation. Middle: 5$^{th}$ percentile volumetric measurement (yellow), contained within the 50$^{th}$ percentile (dark green) and outside that, the 95$^{th}$ percentile (cyan). Right: the probability of a voxel along the black line belonging to the `Whole Tumour' class, as given by different percentiles in the volume space. These percentiles are not yet calibrated, but do contain the ground truth.}
	\label{fig:heuristic}
\end{figure}

We first compute and visualise the predictive uncertainties and segmentations over 20 samples of the model. In Figure \ref{fig:brains}, we display the mean predicted segmentation, and the standard deviation of these outputs. Regions of high predictive variance are common at the borders of segmentation classes. From left to right, with the models becoming more complicated, we see that predicted variance becomes more localised and concentrated around the class boundaries. Empirically, we have found that the model with dropout in every layer produced the most stable variance estimates, whilst being computationally simple to implement. The mean Dice score, for the `whole' label was HR$_{drop\_all}$: 0.86, HR$_{hetero}$: 0.84, HR$_{drop\_last}$: 0.83 and $HR_{\text{default}}$: 0.79. These numbers are average for this task (although the point of this paper is not to win the challenge).
For HR$_{hetero}$, the scale of the aleatoric variance was orders of magnitude lower than the predictive uncertainty and it made no measurable contribution to the overall uncertainty (while being a more expensive model). Because HR$_{drop\_all}$ had the highest Dice scores on average and the aleatoric uncertainty was negligible, we used HR$_{drop\_all}$ for all further results.

In Figure~\ref{fig:heuristic} we plot the value of the probabilistic estimate in the 5$^{th}$ and 95$^{th}$ percentile, produced using the technique in Section~\ref{sec:seguncert}. The volumes form a cumulative distribution function (CDF) over the percentiles. This image provides intuition for our choice to spatially correlate the voxels. The confidence intervals here encompass the ground truth.

\begin{figure}[t]
	\centering
	\begin{subfigure}[t]{3.5cm}
		\centering
		\includegraphics[width=\textwidth]{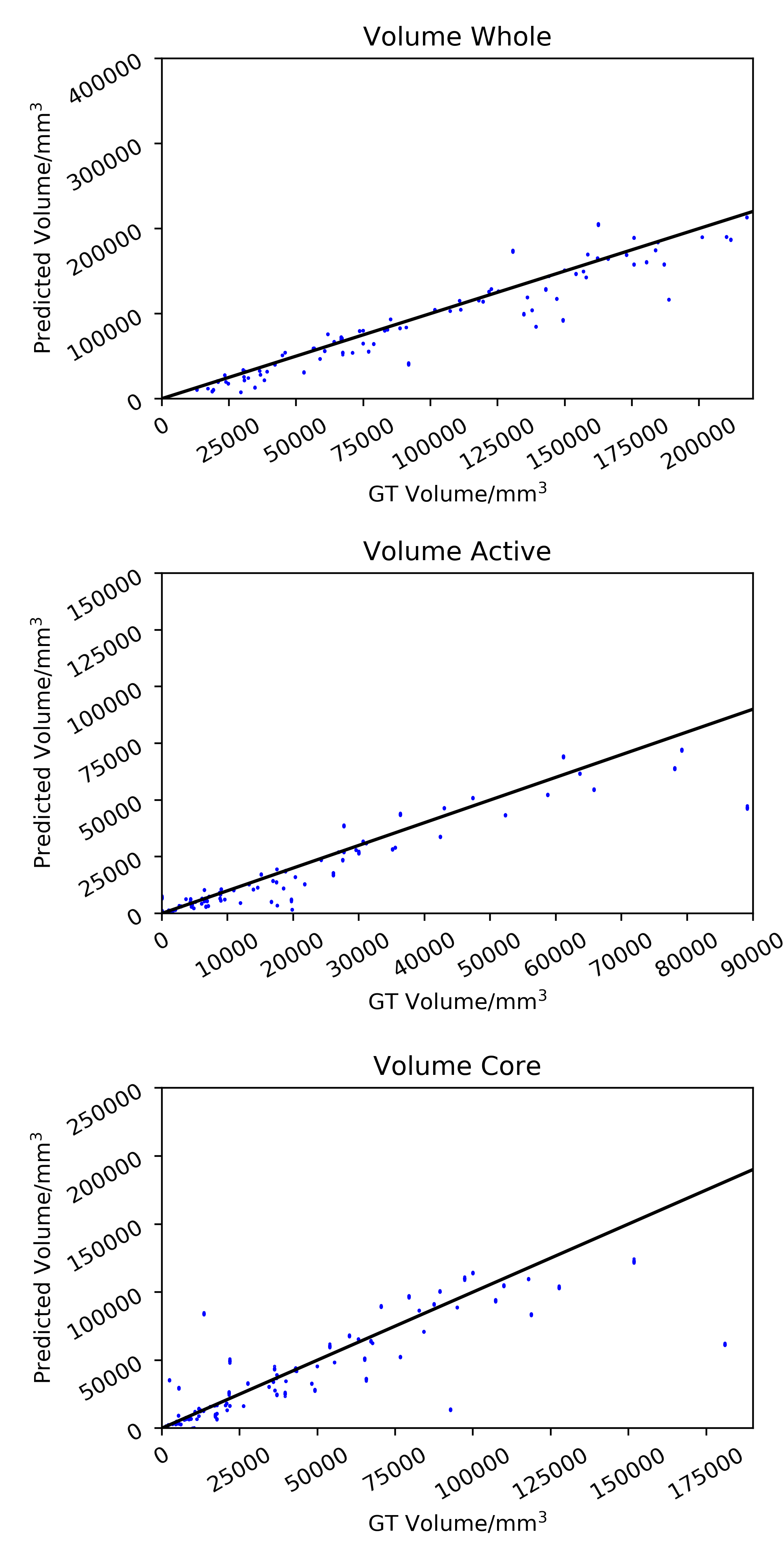}
		\caption{Volume Estimates}\label{fig:1a}		
	\end{subfigure}
	\quad
	\begin{subfigure}[t]{3.5cm}
		\centering
		\includegraphics[width=\textwidth]{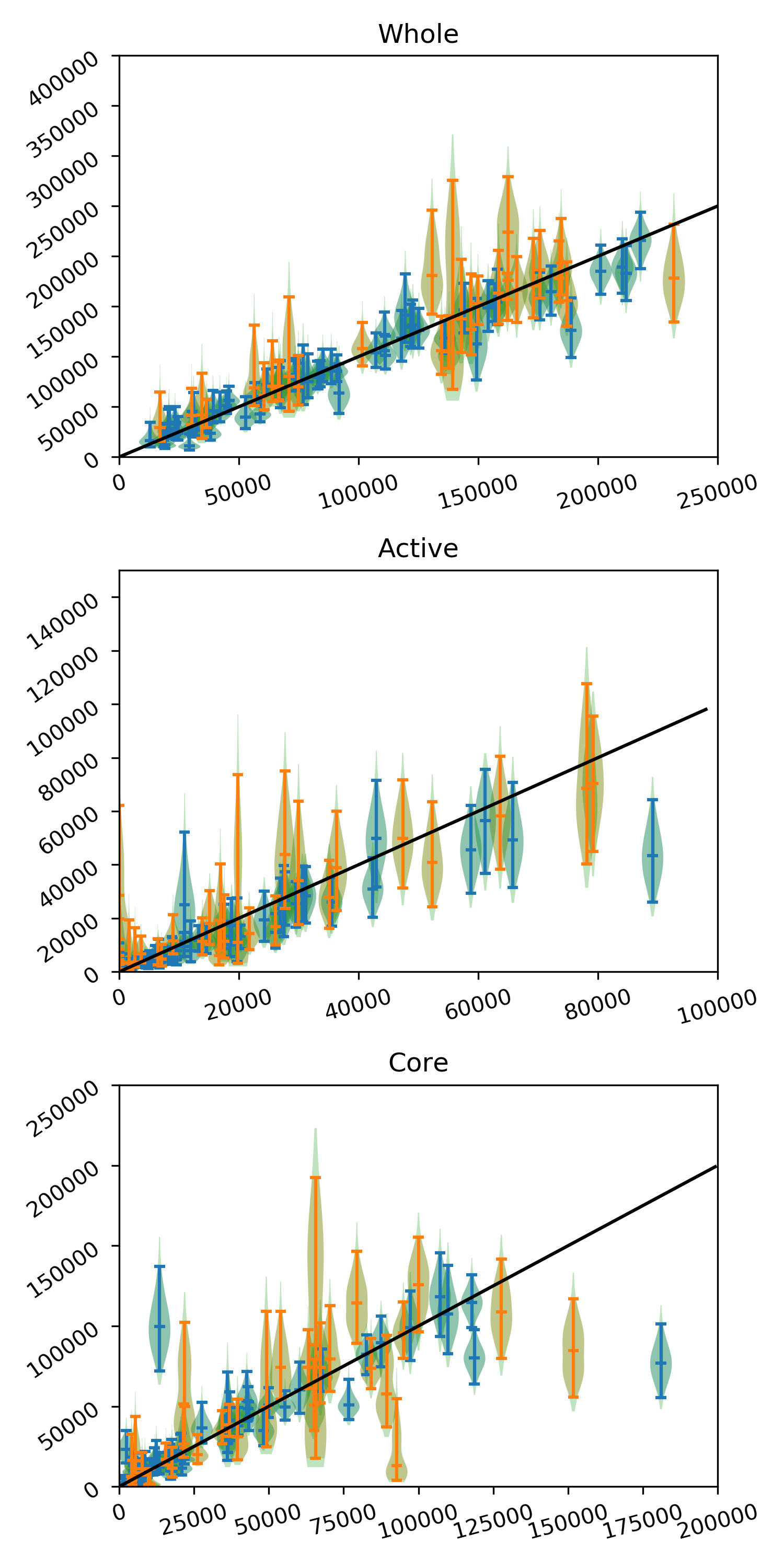}
		\caption{Uncalibrated}\label{fig:1b}
	\end{subfigure}
	\quad
   	\begin{subfigure}[t]{3.5cm}
		\centering
		\includegraphics[width=\textwidth]{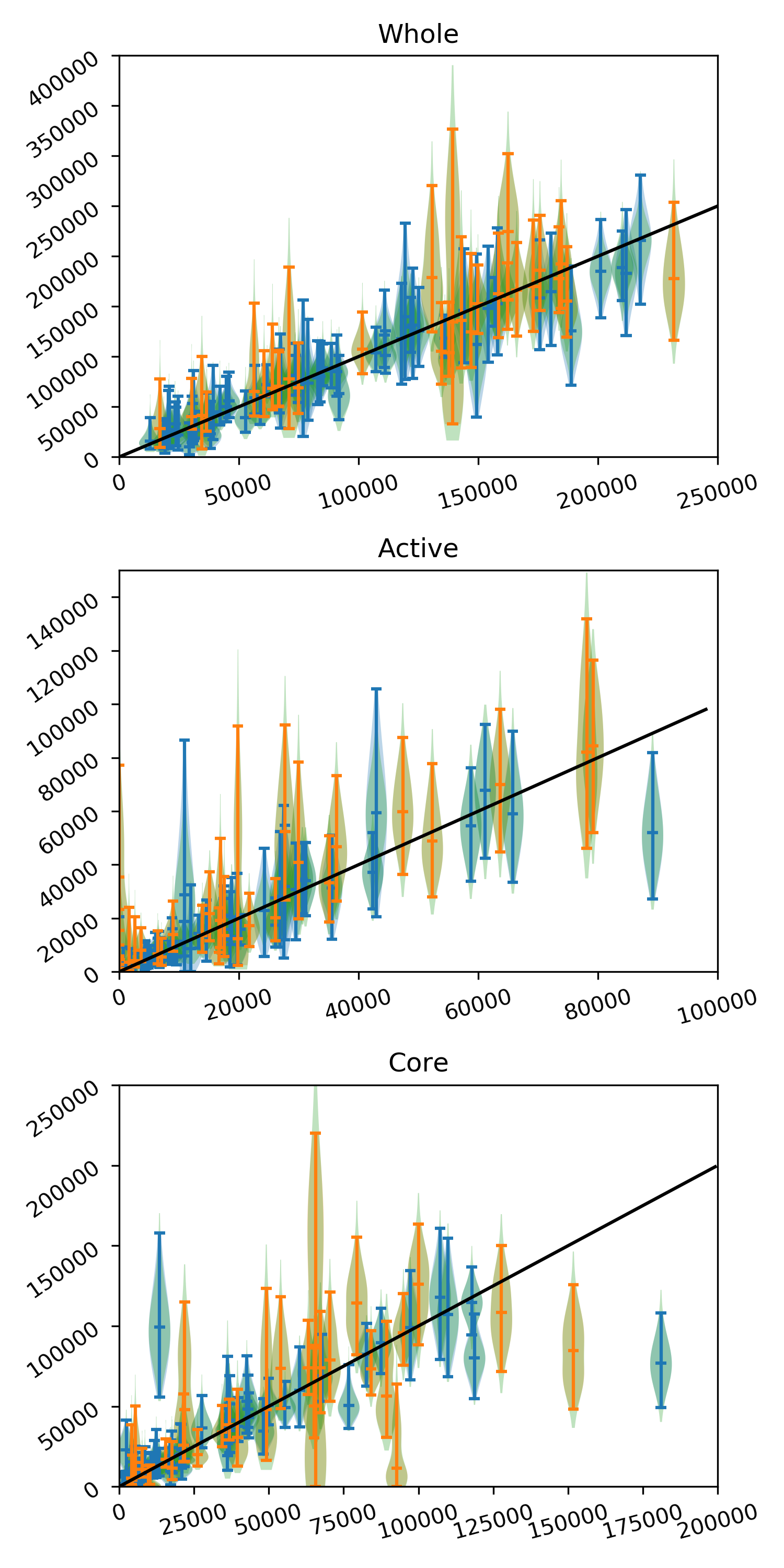}
		\caption{Calibrated}\label{fig:1c}
	\end{subfigure}
	\caption{In Figure \ref{fig:1a}, we plot point estimates for 20 estimates of the volume from HR$_{drop\_all}$. Despite 20 estimates being  plotted, there is almost no variation that can be seen in the plots. This lack of variation is present in all methods (unshown). In \ref{fig:1b}, we see the uncalibrated estimates of uncertainty, and in \ref{fig:1c} we plot the calibrated estimates. Subjects tion' set are blue, and `test' orange. }
	\label{fig:Volumes}
\end{figure}

After fitting the networks, we compare their results in Figure \ref{fig:Volumes}. 
In Figure~\ref{fig:1a}, we plot the predicted volume against the true volume for each of the test subjects. While volumes perform well, with observations clustered around equality 
there is a pronounced lack of variation in the predictions (it is actually impossible to see that 20 points are plotted per subject). This is true for all fitted networks. 
It is immediately apparent that the multiple predictions from any presented modelare highly correlated, and do not end up producing samples with enough variation to include the true value (dashed black line).  The range of estimates is not large enough to act as a confidence interval. 

We thus proceed to forming our volumetric CDF. We plot the distribution of each percentile from this distribution in Figure \ref{fig:1b}, with the mean and 95\% confidence interval being plotted marked with horizontal lines. 
This data is inferred from 200 predicted volumes per subject. While the estimated volumes remain the same in expectation, the spread of the predictions is more commensurate with the known ground truth, although it may remain too small. In Figure~\ref{fig:1c} we present the volume confidence intervals after applying our calibration step.
The confidence intervals, especially in the `Core' segmentation, have longer tails and thus overlap with the line of equality more of the time. 
That the affine scaling enlarges the confidence intervals can be seen empirically correcting the distribution for unmodelled sources of uncertainty. 

\begin{figure}[t]
	\centering
	\includegraphics[width=.92\linewidth]{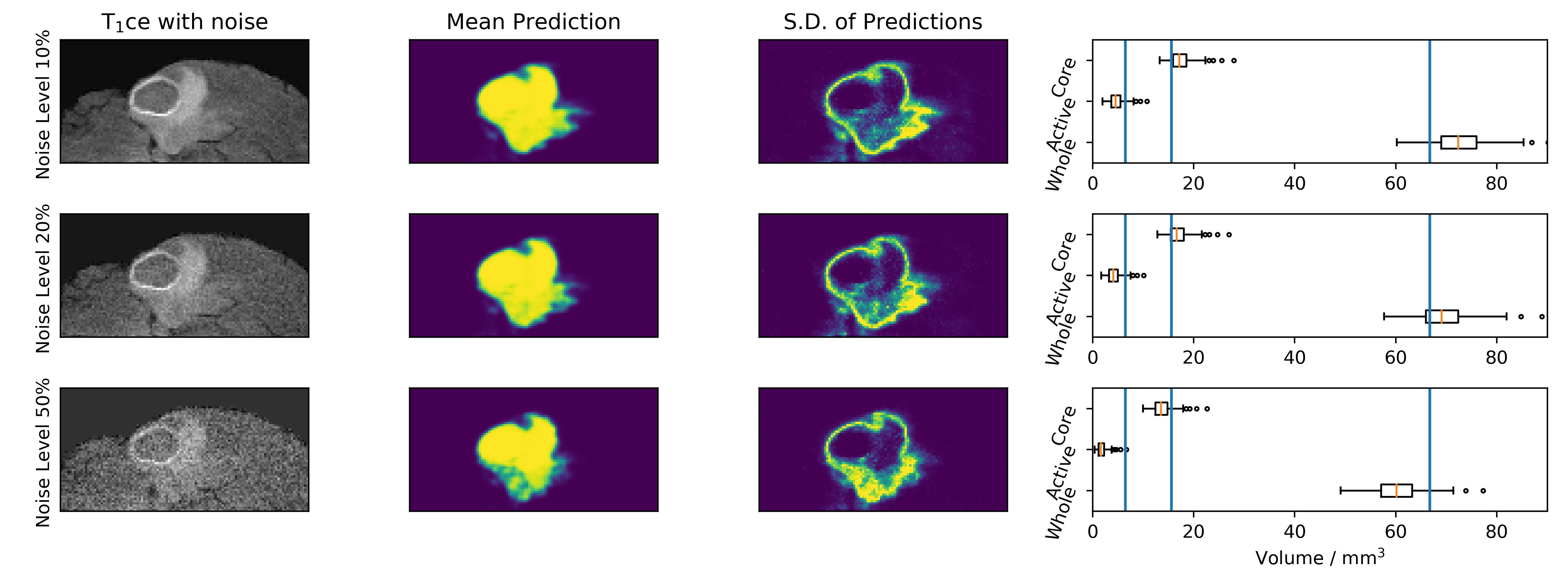}
	\caption{Left column: T$_1$ce image with varying Gaussian noise added, with the $\sigma$ parameter being the labelled percentage of mean foreground intensity. Centre-left: average for `Whole' class for 200 estimates. Centre-right: the predictive uncertainty. Right: volume-level uncertainty, with ground-truth marked blue.  
	}
	\label{fig:noise}
\end{figure}

As a sanity check, we test volume estimation and uncertainty maps on a subject with varying levels of Gaussian noise added to the data in Figure \ref{fig:noise}. As noise increases, we see the segmentation error increase: also, we see the estimated volume reducing with added noise: crucially the confidence intervals still contain the ground truth value. 
One reason that the volume estimate may decrease with added noise is that in regions of uncertainty, cross-entropy loss functions tend to produce the most likely class (in this case, background) --- we did not use weighted cross-entropy. The interaction of this stochastic framework with other loss functions would be an interesting avenue for exploration. While the performance under noise would require more extensive validation, it does raise the possibility of using predicted uncertainty as a proxy for image quality.  

\section{Discussion}

In this work, we have investigated the suitability of single-model estimates of uncertainty in CNNs for producing calibrated confidence intervals of downstream biomarkers. Firstly, we showed these techniques may be applied in medical image segmentation to produce estimates of predictive uncertainty. We also illustrated how measuring volume from different forward passes of the stochastic network was wholly inadequate for producing a range of volumetric estimates that included the ground truth. We then proposed and implemented a solution to calibrate the uncertainty on estimating the volume. 


\textbf{Limitations}: The main limitation of the proposed work is that while we take into account uncertainty in the model parameters and (with HR$_{hetero}$) the data itself, this is not an exhaustive quantification of the error. Factors such as the choice of neural network architecture could also be marginalised over in a Bayesian setting, as alluded to in~\cite{kamnitsas2017ensembles}. 
The empirical nature of calibrating the probabilities is another limitation --- in some settings, there may not be enough withheld data to properly perform the calibration. The calibration step is also, like many machine learning techniques, sensitive to domain shift and relies on continuing statistical similarity between images in the training set and the images for which the application will be used. 

Further work will focus on validating this approach with other medical imaging data. We will investigate extending the proposed methodology for other biomarkers: for instance, uncertainty in shape parameters, or estimates of counts (e.g. for MS lesions). We will also focus on explicitly modeling the effect of network parameter choices on  measured uncertainty (for example, by employing a diverse ensemble of architectures). 

In conclusion, we have shown how to produce calibrated confidence intervals for volumetric analysis, with a non-disruptive extension to a typical deep-learning pipeline. \\
\noindent
\textbf{Acknowledgements:}
ZER is supported by the EPSRC Doctoral Prize. 
FB and MJC are supported by CRUK Accelerator Grant A21993.  
SB is supported by the National Institute for Health Research UCL Biomedical Research Centre.
We gratefully acknowledge NVIDIA Corporation for the donation of hardware. 

\bibliography{bib}

\end{document}